\documentclass[conference]{IEEEtran}
\IEEEoverridecommandlockouts
\usepackage{cite}
\usepackage[T1]{fontenc}
\usepackage{times}
\usepackage{amsmath,amssymb,amsfonts}
\usepackage{algorithmic}
\usepackage{graphicx}
\usepackage{textcomp}
\usepackage{xcolor}
\usepackage{graphicx}
\usepackage{threeparttable}
\usepackage{booktabs}
\usepackage{amsfonts}
\usepackage{subfigure}
\usepackage{amsmath}
\usepackage{threeparttable}
\usepackage{multirow}
\usepackage{paralist}
\usepackage{array}
\usepackage{lineno}
\def\BibTeX{{\rm B\kern-.05em{\sc i\kern-.025em b}\kern-.08em
    T\kern-.1667em\lower.7ex\hbox{E}\kern-.125emX}}

\newcommand*\patchAmsMathEnvironmentForLineno[1]{%
  \expandafter\let\csname old#1\expandafter\endcsname\csname #1\endcsname
  \expandafter\let\csname oldend#1\expandafter\endcsname\csname end#1\endcsname
  \renewenvironment{#1}%
     {\linenomath\csname old#1\endcsname}%
     {\csname oldend#1\endcsname\endlinenomath}}% 
\newcommand*\patchBothAmsMathEnvironmentsForLineno[1]{%
  \patchAmsMathEnvironmentForLineno{#1}%
  \patchAmsMathEnvironmentForLineno{#1*}}%
\AtBeginDocument{%
\patchBothAmsMathEnvironmentsForLineno{equation}%
\patchBothAmsMathEnvironmentsForLineno{align}%
\patchBothAmsMathEnvironmentsForLineno{flalign}%
\patchBothAmsMathEnvironmentsForLineno{alignat}%
\patchBothAmsMathEnvironmentsForLineno{gather}%
\patchBothAmsMathEnvironmentsForLineno{multline}%
}

\usepackage{color}
\newcommand{\blue}[1]{{\color{blue}{#1}}}
\newcommand{\red}[1]{{\color{red}{#1}}}

\def \CD {\mathcal{D}}

\def \CY {\mathcal{Y}}
\def \CT {\mathcal{T}}

\def \R {\mathbb{R}}
\def \CLS {\mbox{CLS}}
\def \SEP {\mbox{SEP}}
\def \BERT {\mbox{BERT}}

\newcolumntype{P}[1]{>{\centering\arraybackslash}p{#1}}

\newtheorem{definition}{Definition}

\begin{document}

\title{Emotion Dynamics Modeling via BERT}

\author{
\IEEEauthorblockN{Haiqin Yang and Jianping Shen} 
\IEEEauthorblockA{Ping An Life Insurance Company of China, Ltd., Shenzhen, China\\
hqyang@ieee.org, shenjianping324@pingan.com.cn
}
}
\maketitle

\begin{abstract}
Emotion dynamics modeling is a significant task in emotion recognition in conversation.  It aims to predict conversational emotions when building empathetic dialogue systems. Existing studies mainly develop models based on Recurrent Neural Networks (RNNs).  They cannot benefit from the power of the recently-developed pre-training strategies for better token representation learning in conversations.  More seriously, it is hard to distinguish the dependency of interlocutors and the emotional influence among interlocutors by simply assembling the features on top of RNNs.  In this paper, we develop a series of BERT-based models to specifically capture the inter-interlocutor and intra-interlocutor dependencies of the conversational emotion dynamics.  Concretely, we first substitute BERT for RNNs to enrich the token representations. Then, a Flat-structured BERT (F-BERT) is applied to link up utterances in a conversation directly, and a Hierarchically-structured BERT (H-BERT) is employed to distinguish the interlocutors when linking up utterances.  More importantly, a Spatial-Temporal-structured BERT, namely ST-BERT, is proposed to further determine the emotional influence among interlocutors.  Finally, we conduct extensive experiments on two popular emotion recognition in conversation benchmark datasets and demonstrate that our proposed models can attain around 5\% and 10\% improvement over the state-of-the-art baselines, respectively.
\end{abstract}


\section{Introduction}
Recently, dialogue systems have achieved significant improvement in many areas thanks to the plethora of publicly available conversational data and the rapid advance of deep learning techniques~\cite{GaoGL19,gao2021advances,lei2020conversational}.  One of the critical challenges to enhancing the systems is generating more human-like conversation~\cite{lei2018sequicity,poria2019emotion,serban2016building}.  Hence, a system should perceive users' emotion states and express the content in an empathetic manner, e.g., by selecting suitable responses from the database or automatically generating human-like responses~\cite{DBLP:journals/tois/HuangZG20,DBLP:journals/inffus/MaNXC20}.  

\begin{figure}[t]
  \label{introduction}
  \centering
  \includegraphics[width=1\columnwidth]{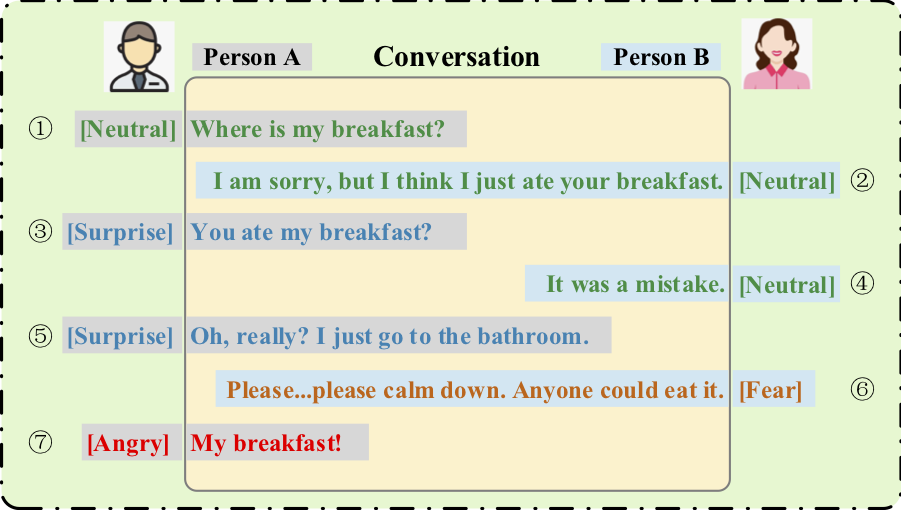}  
  \caption{An example of conversations with emotion dynamics \label{fig:ex}}
\end{figure}

In the literature, Emotion Recognition in Conversation (ERC) is a sub-field of emotion recognition and aims to automatically identify human emotions in conversational scenarios~\cite{abdul2017emonet,poria2019emotion}. A critical task in this field is to model emotion dynamics in conversation~\cite{hazarika2018conversational}.  The emotion dynamics explain the conversational emotion behaviors from two dependencies, i.e., the inter-interlocutor dependency and intra-interlocutor dependency.   In a dialogue, the \textbf{inter}-interlocutor dependency describes the emotional influence \textbf{among} different interlocutors.  That is, one interlocutor tries to coerce other interlocutors changing their emotions~\cite{morris2000emotions}.  For example, as illustrated in Fig.~\ref{fig:ex}, Person B's emotion at timestamp 6 changes from \textit{Neutral} to \textit{Fear} due to the consecutive impact of Person A's strong emotion of \textit{Surprise} at timestamp 3 and 5.  Meanwhile, the \textbf{intra}-interlocutor dependency describes the emotional inertia \textbf{within} individual interlocutors.  That is, one interlocutor tries to resist the change of their own emotion against external influence~\cite{kuppens2010emotional}.  For instance, in Fig.~\ref{fig:ex}, though Person A has changed the emotion to \textit{Surprise} at timestamp 3, Person B remains the original emotion of \textit{Neutral}.

To model emotion dynamics in conversation, researchers have explored various methods based on the Recurrent Neural Networks (RNNs)~\cite{DBLP:conf/naacl/JiaoYKL19,ghosal-etal-2019-dialoguegcn,hazarika2018icon,DBLP:conf/aaai/JiaoLK20,li2020bieru,majumder2019dialoguernn,poria2017context}.  Some studies have adopted the flat structured Recurrent Neural Networks (RNNs), i.e., concatenating utterances of different interlocutors in a single sequence, for context modeling~\cite{li2020bieru,poria2017context}.  Meanwhile, other studies have established variants of hierarchically structured RNNs for context modeling~\cite{DBLP:conf/naacl/JiaoYKL19,ghosal-etal-2019-dialoguegcn,hazarika2018icon,DBLP:conf/aaai/JiaoLK20,majumder2019dialoguernn}, e.g., lining up a sequence of features extracted from a sub-sequence of utterances spoken by the same interlocutor.  However, existing approaches contain the following limitations: (1) the flat structure cannot distinguish the interlocutors because they are blended in the same sequence during temporal modeling; (2) the hierarchical structure cannot distinguish the emotional influence among interlocutors because it applies a flat structure on the extracted sub-sequence features; (3) RNNs can hardly fulfill the power of pre-training language models on large-scale data~\cite{peters2017semi,peters2018ELMo} than recently-developed Transformer-based models~\cite{liu2019roberta,vaswani2017attention,lei2021have}, e.g., BERT~\cite{devlin2019bert}. 

In this paper, we propose a series of BERT-based models to tackle the above challenges.  More specifically, we apply a Flat-structured BERT (F-BERT) to directly link up utterances in a conversation and extend the structure to a Hierarchically-structured BERT  (H-BERT) to distinguish different interlocutors in a conversation.  Both F-BERT and H-BERT can facilitate the power of BERT and can learn better representations by fulfilling pre-training strategies than RNN counterparts.  More importantly, we propose a Spatial-Temporal-structured BERT, namely ST-BERT, to further distinguish emotional influence among interlocutors by first independently capturing the intra-interlocutor and inter-interlocutor emotional influence and then unifying those emotional influence via fusion strategies.

We highlight the contributions of our work as follows: 
\begin{itemize}
\item We develop a series of BERT-based models to explore the potential of applying Transformer-based pre-training models to capture the inter-interlocutor and intra-interlocutor dependencies of the conversational emotion dynamics.  
\item We propose two basic models, F-BERT and H-BERT, as the baselines of their RNN counterparts to exploit the effect of BERT with pre-training strategies.  More importantly, we develop ST-BERT to overcome the weakness of existing work in distinguishing the emotional influence among interlocutors when modeling the emotion dynamics.
\item We conduct extensive experiments on two ERC benchmark datasets to illustrate the effectiveness of our proposed models and observe that we attain  averaged accuracy of about 5\% and 10\% improvement on IEMOCAP and MELD, respectively, over the state-of-the-art (SOTA) baselines.
\end{itemize}

\section{Related Work}

Emotions are hidden mental states associated with human thoughts and feelings~\cite{poria2019emotion}.  Emotion recognition is an interdisciplinary field that spans psychology, cognitive science, machine learning, and natural language processing~\cite{picard2010affective}. The aim is to identify correct emotions from multi-modal expressions~\cite{tsai2019multimodal,tsai2019learning,tzirakis2017end,wang2015constructing}. %Studies on this topic mainly focus on multi-modal fusion.

% have considered all the three modalities, whose primary focus is on fusion strategy while ignoring the Emotion dynamics in a conversation. , where the context plays a vital role

\textbf{Emotion recognition in conversation} (ERC) is to predict the emotion in conversational scenarios. Rather than treating emotions as static states, ERC involves emotion dynamics in a conversation.  By comparing with the recent proposed ERC approaches~\cite{hsu2018emotionlines,majumder2019dialoguernn}, Poria et al.~\cite{poria2019emotion} discover that traditional emotion recognition methods~\cite{colneric2018emotion,kratzwald2018decision,mohammad2010emotions,shaheen2014emotion} fail to perform well because the same utterance within different context may exhibit different emotions.  Mao et al.~\cite{mao2020dialoguetrm} indicate that emotion expressions in different modalities exhibit different dependence on conversational context, where emotion dynamics mainly affect emotion expressions in textual modality.  Recently, various methods have been proposed to tackle ERC in the natural language processing community.  For example, the bi-directional Long Short-Term Memory (LSTM)~\cite{poria2017context} has been applied to capture the intra-interlocutor dependency.  The intra-interlocutor and inter-interlocutor dependencies between dyadic interlocutors have been distinguished by leveraging the hierarchical Gated Recurrent Unit (GRU) and memory networks~\cite{hazarika2018icon,hazarika2018conversational}.  Multiple GRUs with global attention mechanism have been designed and further developed in multi-party ERC~\cite{majumder2019dialoguernn}.  Graph Convolutional Networks (GCNs) have also been employed to mine complex interactions between interlocutors~\cite{ghosal-etal-2019-dialoguegcn,zhang2020generalized}.  GRU-based attention gated hierarchical memory networks have been proposed for ERC~\cite{DBLP:conf/aaai/JiaoLK20}.  However, these methods are mainly based on RNNs and do not sufficiently distinguish the emotional influence among interlocutors.

%Instead of using hierarchical structures, our model adopts a spatial-temporal structure to capture the emotion dynamics in a conversation.

%add BERT Transformer-based model discussion

\textbf{Transformer-based pre-training language models} have applied the Transformer architecture~\cite{vaswani2017attention} to promote language understanding by a two-staged training strategy, i.e., a self-supervised pre-training on a general-domain text corpus and a fine-tune training on the downstream application data.  Pioneers~\cite{peters2017semi,peters2018ELMo} conduct pre-training on bi-directional RNNs to obtain the contextualized representation of each token.  However, RNN-based models are inefficient for long-term modeling and have limited lifting power when stacking more layers due to the limitation of the recurrent connections.  Recently, due to the power of parallel computation and deep-model construction, Transformer-based models, e.g., GPT~\cite{radford2019language}, BERT~\cite{devlin2019bert}, and ELECTRA~\cite{Clark2020ELECTRA}, have been deployed and achieved the SOTA performance in many downstream NLP applications.  Several pieces of work, e.g., transfer learning ERC~\cite{hazarika2019emotion}, utterance-level dialogue understanding~\cite{ghosal2020utterance}, and contextualized emotion sequence tagging~\cite{wang2020contextualized}, have employed pre-training models as a feature extractor in the task of ERC.  However, the potential of the Transformer-based models is less explored and does not address the problem of modeling emotion dynamics yet.

%For our STST, we adopt Transformer-based pre-training models as the fundamental temporal-modeling unit for context modeling in a conversation.

\section{Task Definition}
\label{sec:tf}
We first define the task of utterance-level emotion recognition in conversation.  
\begin{definition}[Emotion Recognition in Conversation]
Let $\CD=\{\CD_i|i\in[1,N]\}$ be a corpus of $N$ conversations, where $\CD_i=\{(u^{\lambda_\tau}_\tau, y_\tau)|\tau\in[1,{L_i}],\lambda_\tau\in[1,S]\}$ is the $i$-th conversation consisting of a sequence of $L_i$ utterances.   $u^{\lambda_\tau}_\tau=w_1\cdots w_{T_\tau}$ is the $\tau$-th utterance of $T_\tau$ words spoken by the ${\lambda_\tau}$-th interlocutor from one of the total $S$ interlocutors. The corresponding emotion type is $y_\tau\in\CY$, where the emotion set $\CY$ consists of all emotions, such as anger, joy, and neutral. The goal is to train a model that can tag each utterance in a new conversation with a emotion label as accurately as possible.
\end{definition}   

%To tackle the above task, we propose ST-BERT to capture different types of emotional influence, for which we need some preliminary notions.
In our work, we aim to capture emotion dynamics in conversations and need the following preliminary notions:
\begin{definition}[Contexts\footnote{In this paper, the notion of context denotes the preceding utterances of the target in a conversation.} in Conversation]
Let $u^{\lambda_i}_i$ be the target utterance in a conversation session 
$U=\{ u^{\lambda_\tau}_\tau|\tau\in[1,L],{\lambda_\tau}\in[1,S] \}$. According to the interlocutors that are involved, we define three types of context utterances within a sliding window of $K$ as follows:
\begin{compactitem}
\item {\bf intra-context:} the preceding utterances of the $i$-th utterance from the interlocutor $\lambda_i$ in $U$ within the window size of $K$: 
\[
\varphi(u^{\lambda_i}_i,U,K)=\{u^{\lambda_\tau}_\tau|\tau\in[\max(i-K, 1), i),{\lambda_\tau}= \lambda_i\},
\]
\item {\bf inter-context:} the preceding utterances of the $i$-th utterance from the interlocutors, except $\lambda_i$, in $U$ within the window size of $K$:  
\[
\phi(u^{\lambda_i}_i,U,K)=\{u^{\lambda_\tau}_\tau|\tau\in[\max(i-K, 1), i),{\lambda_\tau}\neq \lambda_i\}
\] 
\item {\bf conv-context:} the preceding utterances of the $i$-th utterance in $U$ within the window size of $K$:  
\[
\psi(u^{\lambda_i}_i,U,K)=\{u^{\lambda_\tau}_\tau|\tau\in[\max\{i-K,1\},i),{\lambda_\tau}\in [1,S]\}.
\]
\end{compactitem}
\end{definition}

Table~\ref{tab:example} presents an example of the three types of contexts in a conversation. % We try to predict the emotion of $u^{\lambda_i}_i$ by exploiting the context information. %The objective is to predict the emotion of $u^{\lambda_i}_i$ given context information.

\begin{table}[h]
\scriptsize
\centering
  \caption{An example of context utterances in a conversation when $L=8$, $S=3$, and $K=5$. %conversation rounds $L=8$, interlocuter number $S=3$.
  }
  \small
  \label{tab:example}
  \begin{tabular}{p{3cm}|p{4cm}}
    \toprule
    conversation                       & $U= u^1_1\, u^2_2\,u^1_3\,u^1_4\,u^3_5\,u^2_6\,u^1_7\,u^2_8$\\
    \midrule
    target utterance                   & $u^1_7$\\
    \midrule
    intra-context                      & $\varphi(u^1_7,U,5)=u^1_3\,u^1_4$\\
    \midrule
    inter-context                      & $\phi(u^1_7,U,5)=u^2_2\,u^3_5\,u^2_6$\\
    \midrule
    conv-context                       & $\psi(u^1_7,U,5)=u^2_2\,u^1_3\,u^1_4\,u^3_5\,u^2_6$\\
    \bottomrule
  \end{tabular}
\end{table}

\section{BERT}
\begin{figure}[h]
  \centering
  \includegraphics[width=1\columnwidth]{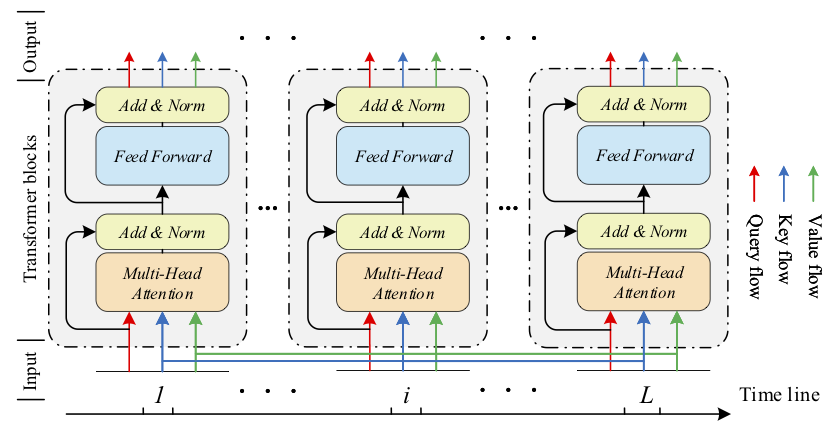}  
  \caption{Architecture of the temporally unfolded Transformer blocks\label{fig:TB}}
\end{figure}
%BERT$_{\mbox{Base}}$ consists of a stack of $N=12$ identical Transformer blocks, which can be expanded in a sequence mode as illustrated in Fig.~\ref{fig:TB} to naturally exploit the temporal information.  
BERT is a powerful Transform-based language model and performs exceptionally well in many downstream NLP tasks~\cite{devlin2019bert}.  A standard BERT model consists of a stack of $L$ identical Transformer blocks expanding in a sequence mode as illustrated in Fig.~\ref{fig:TB} to naturally exploit the temporal information.  Each Transformer block consists of two sublayers, i.e., a multi-head self-attention sublayer and a position-wise fully connected feed-forward sublayer.  Residual connection with layer normalization is employed on each of the two sublayers to avoid the problem of gradient vanishing in training.  The final output is then computed by $\mbox{LayerNorm}(x + \mbox{Sublayer}(x))$, where $\mbox{Sublayer}(x)$ is one of the above mentioned sublayers~\cite{vaswani2017attention}.  In the following, we briefly elucidate the three main layers in BERT.

% and learnable segment embeddings tokenizing the sequence to . 3) . 4) . 

\textit{Embedding layer:} given utterance-context pairs, BERT has 5 key operations in dealing with the input pairs: (1) packing the utterance and its context into a sequence of tokens by WordPiece tokenization; (2) adding $[\CLS]$ as the classification token at the head of a sequence; (3) adding the special token $[\SEP]$ and token type embeddings to differentiate the utterance and its context; (4) applying WordPiece embeddings~\cite{WuSCLNMKCGMKSJL16} on the tokens; (5) adding position embeddings to maintain the order information in a sequence. 

\textit{Multi-head attention sublayer:} the output of embedding layer is fed to the multi-head self-attention sublayer (the orange blocks in Fig.~\ref{fig:TB}) to model temporal dependency.  A multi-head attention independently applies attention mechanism $H$ times along with the query $Q\in\R^{d_k}$, key $K\in\R^{d_k}$, and value $V\in\R^{d_v}$, respectively, and concatenates them together:
\begin{align}
& \mbox{MultiHead}(Q, K, V) = \mbox{Concat}(head_1,...,head_H)W^O,\\\nonumber
& head_h = \mbox{Attention}(QW_h^Q, KW_h^K, VW_h^V), \\\nonumber
&\qquad~~ = \mbox{softmax}(\frac{QW_h^Q(KW_h^K)^T}{\sqrt[]{d_k}})VW_h^V,~~ h=1,\ldots, H,
\end{align}
where $\frac{1}{\sqrt{d_k}}$ is a scaling factor to avoid pushing the softmax function into small gradients regions when $d_k$ is large~\cite{vaswani2017attention}.  $head_h$ is the $h$-th output of attention operation.  $W_h^Q\in\R^{d_{model}\times d_k}, W_h^K\in\R^{d_{model}\times d_k}, W_h^V\in\R^{d_{model}\times d_v},W^O\in\R^{Hd_v\times d_{model}}$ are the weights of the projection matrices for computing the $h$-th attention operations.  Practically, we set the dimension $d_k=d_v=d_{model}/H$.
%where $head_h = \mbox{Attention}(QW_h^Q, KW_h^K, VW_h^V) = \mbox{softmax}(\frac{QW_h^Q(KW_h^K)^T}{\sqrt[]{d_k}})VW_h^V$, $h=1,\ldots, H$.  The constant $\frac{1}{\sqrt[]{d_k}}$ is a scaling factor to avoid pushing the softmax function into small gradients regions when $d_k$ is large~\cite{vaswani2017attention}.   $head_h$ is the $h$-th output of attention operation.  $W_h^Q\in\R^{d_{model}\times d_k}, W_h^K\in\R^{d_{model}\times d_k}, W_h^V\in\R^{d_{model}\times d_v},W^O\in\R^{Hd_v\times d_{model}}$ are the weights of the projection matrices for computing the $h$-th attention operations.  Practically, we set $d_k=d_v=d_{model}/H$.
\if 0
\begin{align}
head_h &= \mbox{Attention}(QW_h^Q, KW_h^K, VW_h^V)\\
       &= \mbox{softmax}(\frac{QW_h^Q(KW_h^K)^T}{\sqrt[]{d_k}})VW_h^V
\end{align}
\fi 

\textit{Fully connected feed-forward sublayer:} after applying the multi-head attention, a two-layer fully connected feed-forward network is computed by a RELU on the hidden state:
\begin{equation}
    \max(0, xW_1 + b_1)W_2 + b_2,
\end{equation} % $$, 
where $x$ is the output from the multi-head attention sublayer, $W_1\in\R^{d_{model}\times d_{hidden}}$, $W_2\in\R^{d_{hidden}\times d_{model}}$, and $b_1\in\R^{d_{hidden}}$, $b_2\in\R^{d_{model}}$ are the projection matrices and biases for the fully-connected networks, respectively. 

%\textit{Residual connection with layer normalization:} The output of each sublayer is surrounded by residual connections which can be denoted as:
%\begin{equation}
%Output_{Sublayer}(x) = LayerNorm(x+Sublayer(x))
%\end{equation}
%where $Sublayer$ can be either the multi-head attention or the feed-forward network. $LayerNorm$ is the computation of layer normalization~\cite{ba2016layer}. More details about residual connection can be found in~\cite{he2015deep}.

\section{Our Proposal}
In the following, we present our proposed series of BERT-based models. 
\subsection{F-BERT}
\begin{figure}[t]
  \centering
  \subfigure[Flat Structure]{\includegraphics[width=0.31\columnwidth]{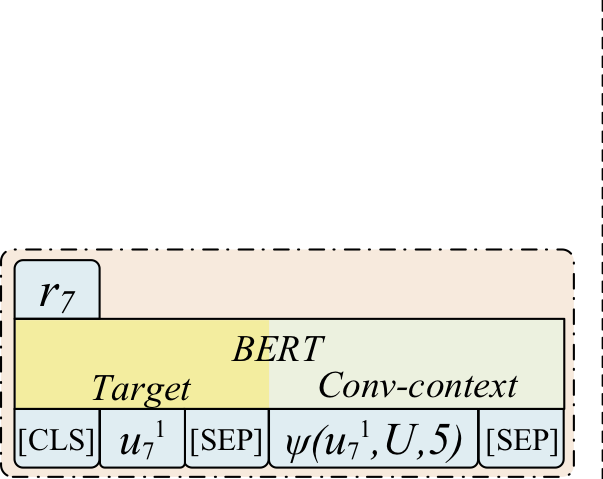}\label{fs}}     
  \subfigure[{Hierarchical Structure}]{\includegraphics[width=0.645\columnwidth]{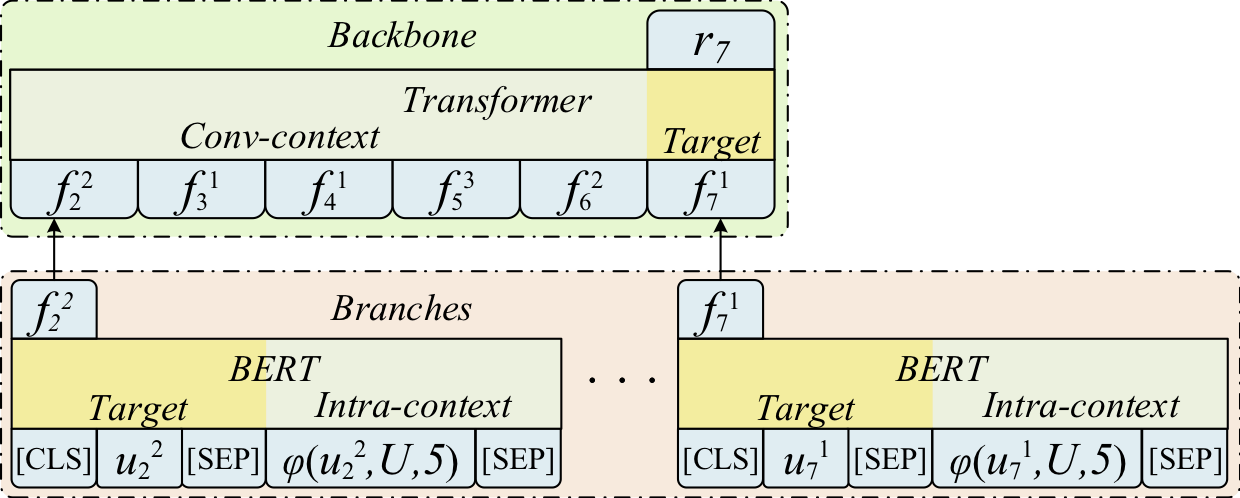}\label{hs}}
  \subfigure[{Spatial-Temporal Structure}]{\includegraphics[width=1\columnwidth]{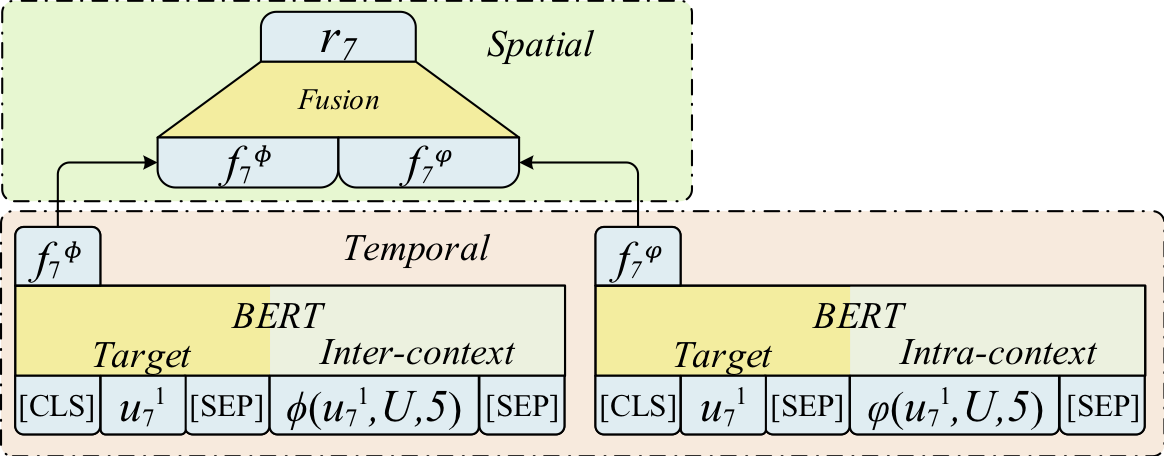}\label{sts} }
  \caption{Architecture of the proposed BERT-based models given inputs of table~\ref{introduction}}
\end{figure}

{The Flat-structured BERT (F-BERT), as illustrated in Fig.~\ref{fs}, is to directly {concatenate} the target utterance with the conv-context while applying BERT afterwards.} The input is a sequence of utterance-context pair, i.e., the {target utterance} of $T$ sub-words, $u^{\lambda_i}_i=w_1\cdots w_{T}$, and the  {conv-context} of $\CT_\psi$ sub-words, $\psi(u^{\lambda_i}_i, U, K)=\omega_1\cdots \omega_{\CT_\psi}$. We pack the pairs into a sequence of tokens:
\begin{equation}
X_i=  [\CLS]\,u^{\lambda_i}_i\,[\SEP]\,\psi(u^{\lambda_i}_i, U, K)\,[\SEP].  
\end{equation}
Hence, $X_i$ includes the information of the target utterance with the conversation context, but does not distinguish the identity of interlocutors.

After pre-processing, $X_i$ is fed to BERT and represented by the last hidden layer at the $[\CLS]$ token, denoted by
\begin{equation}
r_i=\BERT(X_i),
\end{equation}
where $r_i$ is the output representation of $u^{\lambda_i}_i$ for emotional predictions.

\subsection{H-BERT}
\label{sec:HS}

%is constructed by using the BERT as the branches to distinguish interlocutors and using the Transformer as the backbone to temporally wrap all the emotion influence captured by the BERT.}
{The Hierarchically-structured BERT (H-BERT), as illustrated in Fig.~\ref{hs}, first applies BERT to wrap up an utterance with its intra-context to capture the intra-interlocutor dependency as the branch feature.  Next, it lines up the branch features in conversational order by the backbone Transformer to produce the final output representation.}  More specifically, the input of a branch BERT is a sequence of utterance-context pair, i.e., the target utterance of $T$ sub-words, $u^{\lambda_i}_i=w_1\cdots w_T$, and the intra-context of $\CT_\varphi$ sub-words, $\varphi(u^{\lambda_i}_i,U,K)=\omega_1\cdots \omega_{\CT_\varphi}$. After packing, we compute the input sequence by 
\begin{equation}
X^{\lambda_i}_i =  [\CLS]\,u^{\lambda_i}_i\,[\SEP]\,\varphi(u^{\lambda_i}_i,U,K)\,[\SEP].
\end{equation}
Thus, $X^{\lambda_i}_i$ includes the information of the target utterance with historical utterances only spoken by the target interlocutor. By feeding $X^{\lambda_i}_i$ to BERT, we obtain 
\begin{equation}
f^{\lambda_i}_i=\BERT(X^{\lambda_i}_i),
\end{equation}
we can obtain $f^{\lambda_i}_i$, the intra-interlocutor dependency of the $\lambda_i$-th interlocutor, which maintains the $i$-th emotion influence in a conversation.
%we can obtain $f^{\lambda_i}_i$, which maintains the emotion influence at $i$-th utterance and the intra-interlocutor dependency of the $\lambda_i$-th interlocutor. %, which maintains the emotion influence at $i$-th utterance. % of the $\lambda_i$-th interlocutor. 

{Let $F=f^{\lambda_1}_1\cdots f^{\lambda_L}_L$ be the sequence of the branch features of the entire conversation. The input of the backbone Transformer is a sliding window of $K+1$ in $F$: %, which is
\begin{equation}
F_i =f^{\lambda_{i-K}}_{i-K}\cdots f^{\lambda_{i-1}}_{i-1}f^{\lambda_i}_i.
\end{equation}
where $f^{\lambda_i}_i$ is the target feature at the last position.  $f^{\lambda_{i-K}}_{i-K}\cdots f^{\lambda_{i-1}}_{i-1}$ are the preceding $K$ features that may produce emotional influence to the target.  $F_i$ is fed to the backbone Transformer and represented by the last hidden layer at the target position.  The computation of the Transformer is
\begin{equation}
r_i=\mbox{Transformer}(F_i),
\end{equation}
where $r_i$ is the representation of $u^{\lambda_i}_i$ conditioned on the temporally wrapped emotion influence for making predictions.}

\subsection{ST-BERT}

The Spatial-Temporal-structured BERT (ST-BERT), as illustrated in Fig.~\ref{sts}, first applies BERT to individually wrap up the inter-context and intra-context of the target utterance as two types of temporally captured emotion influence.  After that, different fusion strategies are applied to combine the two temporally captured emotional influence into a single spatial representation.
%for corresponding emotion influence in the temporal dimension and using the fusion module to combine those two emotional influences in the spatial dimension. The combined representation is for emotional predictions.

\textbf{Temporal Construction.} 
Given the target utterance $u^{\lambda_i}_i=w_1\cdots w_T$ of $T$ sub-words, we construct two individual sequences of utterance-context pairs: 
\begin{align}
& X^\varphi_i = [\CLS]\,u^{\lambda_i}_i\,[\SEP]\,\varphi(u^{\lambda_i}_i,U,K)\,[\SEP],\\
%\end{equation}
%\begin{equation}
& X^\phi_i = [\CLS]\,u^{\lambda_i}_i\,[\SEP]\,\phi(u^{\lambda_i}_i,U,K)\,[\SEP],
\end{align}
where $X^\varphi_i$ includes the information, which maintains the emotion inertia of the $\lambda_i$-th interlocutor by incorporating the intra-context $\varphi(u^{\lambda_i}_i,U,K)=\omega^\varphi_1\cdots \omega^\varphi_{\CT_\varphi}$. $X^\phi_i$ includes the information, which produces emotional influence from the non-$\lambda_i$ interlocutor by incorporating the inter-context $\phi(u^{\lambda_i}_i,U,K)=\omega^\phi_1\cdots \omega^\phi_{\CT_\phi}$. The two types of token sequences are then fed to the BERT, 
\begin{equation}
f^\varphi_i=\BERT(X^\varphi_i), \quad
%\end{equation}
%\begin{equation}
f^\phi_i=\BERT(X^\phi_i ),
\end{equation}
we can directly and explicitly capture the intra-interlocutor and inter-interlocutor dependencies in $f^\varphi_i\in\R^{d_{model}}$ and $f^\phi_i\in\R^{d_{model}}$, respectively. The two temporally contextualized features are fused vertically in different spatial fusion strategies.   %fed to the spatial dimension for feature fusion. % combination.

\textbf{Spatial Fusion.} 
We explore three types of spatial fusion strategies, including direct concatenation, gate operation, and attention mechanism.  
%linear weighting, % approaches like concatenation (a linear weighting strategy), gate (a neuron-level interactive weighting strategy), and attention (a vector-level interactive weighting strategy). 

\textit{Direct concatenation} is a simple but effective strategy.  It does not involve interactions between features and can be computed by a fully-connected network: 
\begin{equation}
r_i=W^C[f^\varphi_i; f^\phi_i] + b^C,
\end{equation}
where $[\cdot;\cdot]$ is the concatenation. $r_i\in\R^{d_{hidden}}$ is the output representation for emotional predictions. $W^C$ and $b^C$ are the projection matrix and bias, respectively.

\textit{Gate operation} is a neuron-level interactive weighting strategy.  The computation can be formulated as
\begin{align}
& h^\varphi_i=\tanh(W^\varphi f^\varphi_i + b^\varphi),\\
%\end{equation}
%\begin{equation}
&h^\phi_i=\tanh(W^\phi f^\phi_i + b^\phi),\\
%\end{equation}
%\begin{equation}
&z_i=\sigma(W^Z [f^\varphi_i; f^\phi_i] + b^Z),\\
%\end{equation}
%\begin{equation}
&r_i=z_i * h^\varphi_i + (1-z_i)*h^\phi_i,
\end{align}
where $h^\varphi_i\in\R^{d_{hidden}}$ and $h^\phi_i\in\R^{d_{hidden}}$ are the projections of $f^\varphi_i$ and $f^\phi_i$, respectively. $*$ refers to the Hadamard product whose function is to use neurons in one vector to weight the neurons of its counterpart at the same position. Here, $z_i\in\R^{d_{hidden}}$ is to weight neurons in $h^\varphi_i$ and $1-z_i$ is to weight neurons in $h^\phi_i$, where $z_i$ is computed by feature interactions between $f^\varphi_i$ and $f^\phi_i$, $\sigma$ is to maps the $z_i$ to $(0,1)$. Note that the impact of ``$1-z_i$" on $h^\phi_i$ is critical  because it allows to learn the weights of the neurons in $h^\varphi_i$ and $h^\phi_i$ contrastively~\cite{arevalo2020gated}.  $W^\varphi$ and $b^\varphi$, $W^\phi$ and $b^\phi$, and $W^Z$ and $b^Z$, are the corresponding projection matrices and biases, respectively.  $r_i$ is the final output representation.   

\textit{Attention mechanism} is a vector-level interactive weighting strategy.  The insight is identical to that of the gate operation. The major difference is that gate operation allocate different weights to neurons in a vector while attention operation allocate the same weight to neurons in a vector.  The attention operation can be computed by
\begin{align}
&\alpha_i=\sigma(\frac{f^\varphi_i (f^\phi_i)^T}{\sqrt{d_{model}}}),\\
%\end{equation}
%\begin{equation}
\label{eq:ST-BERT-att}
&r_i=\alpha_i * h^\varphi_i + (1-\alpha_i)*h^\phi_i,
\end{align}
where $\frac{1}{\sqrt[]{d_{model}}}$ is the scaling factor. $d_{model}$ is the dimention of  $f^\varphi_i, f^\phi_i$. $\alpha_i$ is the attention weight and a scalar computed by dot-product between $f^\varphi_i$ and $f^\phi_i$.  Again, the operation of ``$1-\alpha_i$" plays a trade-off on the weight for contrastive learning~\cite{cho2014learning}. $r_i$ is the final output representation.

\subsection{Discriminator}
The discriminator is a two-layer perceptron with the hidden layer activated by the $\tanh$ function, which can be trained by minimizing the cross-entropy loss.  Given the representation $r_i$, the discriminator output emotion distributions computed by the softmax function: %. The emotion type of the highest probability is the predicted emotion. The formulation of the discriminator is
\begin{align}
&o_i=\tanh(W^O \cdot r_i),\\
%\end{equation}
%\begin{equation}
&\mathcal{P}_i=\mbox{softmax}(W^{\mathcal{P}_i}\cdot o_i),\\
%\end{equation}
%\begin{equation}
&\hat{y}_i=\mathop{\arg \max}\limits_k\mathcal{P}_i[k],
\end{align}
where $\hat{y}_i$ is the predicted emotion. $W^O$ and $W^{\mathcal{P}_i}$ are the corresponding projection matrices. %We use cross-entropy loss to train the model.

\section{Experiments}
In this section, we present the experiments with detailed analysis.
\subsection{Datasets}
Two popular ERC datasets, IEMOCAP~\cite{busso2008iemocap} and MELD~\cite{poria2019meld}, are adopted to evaluate our proposed models. Statistics on the two datasets are presented in Table~\ref{fig:stat}.

{IEMOCAP} consists of dyadic conversation videos between pairs of 10 speakers. Each utterance is annotated with one of the six emotion types, including happy, sad, neutral, angry, excited, and frustrated. Following~\cite{majumder2019dialoguernn}, we apply the first four sessions in training and use the last session for test.  The validation conversations are randomly selected from the training set with a ratio of 0.1.

{MELD} consists of multi-party conversation videos collected from the Friends TV series. Each utterance is annotated with one of the seven emotion types, including anger, disgust, sadness, joy, neutral, surprise, and fear. We apply the official splits for training, validation, and test in this dataset.

% \textit{DailyDialogue} consists of dyadic human-written conversations covering various topics about human daily life. Each utterance is annotated with one of the seven emotion types, including anger, disgust, fear, happiness, sadness, surprise, and neutral. DailyDialogue provides official splits for training, validation, and testing. We follow~\cite{wang2020contextualized} that excludes neutral utterance due to the imbalanced data distribution (more than 80\% of the utterances are neutral).  Since the dataset does not provide speaker information, we assume all the conversations are based on the ``ABAB...'' pattern. 

\begin{table}[htp]
\centering
\caption{Statistics of the two datasets}
\label{fig:stat}
\begin{tabular}{l|P{20pt}|P{20pt}|P{20pt}|P{25pt}|c}
\toprule
\multirow{2}{*}{Dataset} & \multicolumn{3}{c|}{\# Utterances} & \multirow{2}{*}{Classes} & \multirow{2}{*}{Avg. conv. length} \\
                         & train      & val       & test     &                           &                                           \\
\midrule
IEMOCAP                  & 5,663      & 647       & 1,623    & 6                         & 50                                        \\
MELD                     & 9,989      & 1,109     & 2,610    & 7                         & 8                                         \\
% DailyDialogue            & 11,118     & 1,000    & 1,000    & 87,179     & 8,069     & 7,740    & 7                         & 10   \\ 
\bottomrule
\end{tabular}
\end{table}

\subsection{Implementation Details}
In the experiment, BERT$_{\tiny\mbox{Base}}$ is adopted as the fundamental sequential module, where it consists of 12 Transformer blocks, 12 self-attention heads, and 768 hidden-units.  We employ the off-the-shelf implementation of BERT$_{\tiny\mbox{Base}}$ model in ``\textit{Transformers}"\footnote{https://github.com/huggingface/transformers}. The pre-trained parameters are deployed for initializing BERT. Other parameters are randomly initialized. All the hyperparameters are keeping default. The backbone of the H-BERT is a 6-layer, 12-head-attention, and 768 hidden-unit Transformer encoder implemented using \textit{torch.nn.TransformerEncoder}\footnote{https://pytorch.org/docs/stable/generated/torch.nn.TransformerEncoder.html} in PyTorch. The parameters of the backbone Transformer are randomly initialized. We use AdamW~\cite{loshchilov2018fixing} as the optimizer with the following setup: an initial learning rate of $6e-6$, $\beta_1=0.9$, $\beta_2=0.999$, L2 weight decay of $0.01$, learning rate warms up steps being $0$, and linear decay of the learning rate. All the results are based on an average of 5 runs. 
%For simplification, the proposed model does not expand to multi-GPU settings. Our hardware affords a maximum conversational context size of 14. A larger context can achieve better performance~\cite{hazarika2018icon}, which is beyond the scope of this paper.

\begin{table*}[htp]
\small
\centering
\begin{threeparttable}
  \caption{Main Results on IEMOCAP and MELD.\label{tab:rs}}
  \begin{tabular}{l|p{16pt}p{16pt}|p{16pt}p{16pt}|p{16pt}p{16pt}|p{16pt}p{16pt}|p{16pt}p{16pt}|p{16pt}p{16pt}|p{16pt}p{16pt}|p{16pt}p{16pt}}
    \toprule
    \multirow{3}{*}{Model} & \multicolumn{14}{c|}{IEMOCAP} & \multicolumn{2}{c}{MELD}\\
    \cline{2-17}
    & \multicolumn{2}{c|}{happy} & \multicolumn{2}{c|}{sad} & \multicolumn{2}{c|}{neutral} & \multicolumn{2}{c|}{angry} & \multicolumn{2}{c|}{excited} & \multicolumn{2}{c|}{frustrated} & \multicolumn{2}{c|}{\textbf{Avg.(w)}} & \multicolumn{2}{c}{\textbf{Avg.(w)}}\\
                 & ACC   & F1    & ACC   & F1    & ACC   & F1    & ACC   & F1    & ACC   & F1    & ACC   & F1    & ACC   & F1 & ACC & F1        \\
    \midrule\midrule
    scLSTM       & 37.5  & 43.4  & 67.7  & 69.8  & 64.2  & 55.8  & 61.9  & 61.8  & 51.8  & 59.3  & 61.3  & 60.2  & 59.2  & 59.1& 57.5  & 55.9 \\
    TL-ERC       & -    & -     & -     & -     & -     & -     & -     & -     & -     & -     & -     & -     & -     & 58.8& -        & -    \\  
    %CMN          & 25   & 30.4 & 55.9   & 62.4 & 52.9 & 52.4 & 61.8 & 59.8 & 55.5 & 60.3 & 71.1 & 60.7 & 56.6 & 56.1& -        & -    \\
    %ICON         & 23.6 & 32.8  & 70.6  & 74.4  & 59.9  & 60.6  & 68.2 & \textbf{68.2} & 72.2  & 68.4  & \textbf{71.9}  & 66.2 & 64.0 & 63.5& -       & -    \\
    DRNN*        & 25.7 & 33.2 & 75.1   & 78.8  & 58.6 & 59.2 & 64.7 & 65.3 & \textbf{80.3} & 71.9 & 61.2 & 59.0 & 63.4  & 62.8& 56.1  & 55.9 \\
    DRNN$\dag$*  & - & - & -   & - & - & - & - & -& - & - & - & - & - & 64.1  & -        & - \\
    DGCN*        & 40.6 & 42.8 & \textbf{89.1}   & \textbf{84.5} & 61.9 & 63.5 & 67.5 & 64.2 & 65.5 & 63.1 & 64.2 & \textbf{67.0} & 65.3 & 64.2  & -        & 58.1 \\
    
    AGHMN        & 48.3 & \textbf{52.1} & 68.3   & 73.3  & 61.6  & 58.4  & 57.5  & 61.9  & 68.1  & 69.7  & 67.1  & 62.3  & 63.5  & 63.5& 59.5  & 57.5 \\
    CESTa*       & -    & 47.7 & -      & 80.8  & -    & 64.8 & -    & 63.4 & -    & \textbf{76.0} & -   & 62.7 & - & 67.1 & -      & 58.4 \\
    \midrule\midrule
    F-BERT       & \textbf{52.8} & 50.8 & 77.6 & 77.6 & 65.1 & 64.1 & \textbf{71.2} & 61.9 & 65.6 & 69.4 & 58.5 & 61.9 & 65.1 & 65.2 & 63.0 & 62.5\\
    H-BERT       & 30.8 & 33.2 & 68.4 & 73.9 & 69.2 & 68.7 & 66.0 & 59.0 & 70.3 & 70.0 & 65.1  & 64.6 & 65.1 & 65.1 & 63.2 & 62.6\\
    \midrule\midrule
    ST-BERT-ATT  & 37.5 & 41.7 & 80.0 & 77.2 & 72.1 & 66.3 & 54.7 & 57.6 & 63.2 & 68.6 & 64.6 & 64.0 & 65.0 & 64.7 & 64.4 & 62.7 \\
    ST-BERT-CON  & 46.5 & 49.1 &81.2 & 80.4 &\textbf{76.3} & \textbf{68.6}& 61.2 & 62.7 & 68.2 & 71.7& 60.4 & 63.6 & 67.6 & 67.4 & 65.5 & 63.5 \\
    ST-BERT-GAT  & 45.1 & 48.5 & 80.0 & 82.2 & 72.4 & 68.6 & 66.5 & \textbf{65.6} & 66.9 & 69.9 & \textbf{67.2} & 66.6 & \textbf{68.5} & \textbf{68.4} & \textbf{65.7} & \textbf{63.8} \\
    \bottomrule 
  \end{tabular}
  \begin{tablenotes}
      \scriptsize
      \item Symbol * indicate that models are fed with succeeding context. DRNN$\dag$*~\cite{ghosal2020utterance} is fed with RoBERTa~\cite{liu2019roberta} features. -ATT, -CON, and -GAT denote the attention, concatenation, and gate mechanism in ST-BERT.
  \end{tablenotes}
\end{threeparttable}
\end{table*}

\subsection{Comparing Methods and Metrics}
We investigate previous ERC methods based only on the textual modality:
\begin{compactitem}
\item {scLSTM}~\cite{poria2017context} is the earliest study that we can track in the task of ERC. It makes predictions by only considering intra-interlocutor dependency.

\item {TL-ERC}~\cite{hazarika2019emotion} applies BERT as a transfer learning module for context modeling. That is, it simply employs BERT as a feature extractor and applies RNN for modeling emotion dynamics in conversation afterwards.

%\textit{CMN}~\cite{hazarika2018conversational} is a memory-network-based model that first distinguishes interlocutors for conversation modeling. The results are quoted from ~\cite{hazarika2018icon}.

%\textit{ICON}~\cite{hazarika2018icon} is an extension of CMN. The modification is using another GRU to explicitly model the inter-personal dependency.

\item {DRNN}~\cite{majumder2019dialoguernn} is DialogueRNN, a hierarchical attention-based model with three GRUs to capture the emotion dynamics.  In the experiment, we compare both DRNN with the CNN and DRNN with the RoBERTa features~\cite{liu2019roberta,ghosal2020utterance}, denoted by DRNN$\dag$.

\item {DGCN}~\cite{ghosal-etal-2019-dialoguegcn} applies GCN to model utterance interactions among interlocutors by considering speaker positions in the historical conversation.

\item {AGHMN}~\cite{DBLP:conf/aaai/JiaoLK20} finetunes sentence representation and uses GRU to wrap the attention-weighted representations rather than summing them up.

\item {CESTa}~\cite{wang2020contextualized} is the SOTA ERC method.  It first applies a flat Transformer at the bottom layer to obtain contextualized representation for each utterance in the conversation.  Next, it cascades a hierarchical LSTM upon the Transformer to distinguish contextualized representations of interlocutors.
\end{compactitem}

Following~\cite{ghosal-etal-2019-dialoguegcn,majumder2019dialoguernn}, we use the weighted accuracy (ACC)  and weighted average F1 (F1) as the evaluation metrics:
\begin{align}
    \mbox{ACC} = \sum_{c=1}^{|\mathcal{C}|}p_c\cdot a_c,\quad
    \mbox{F1} &= \sum_{c=1}^{|\mathcal{C}|}p_c\cdot \mbox{F1}_c,
\end{align}
where $p_c$ is the percentage of the class $c$ in the testing set, $a_c$ and $F1_c$ are the corresponding accuracy and F1 score for the class $c$, respectively.  It is worth mentioning that we mainly focus on the average scores because all the methods have trade-off among individual emotion types. 
\subsection{Main Results}
Table~\ref{tab:rs} reports the main results.  We can observe that  
\begin{compactitem}[--]
\item The Transformer-based models, i.e., {CESTa} and our proposed models, achieve better performance than RNN-based models on IEMOCAP.  Notice that, the results of F-BERT and H-BERT are competitive.  By observing the average conversation length on IEMOCAP is 50, we can note that Transformer-based models can effectively capture information for long sentences.
\item The average ACC and F1 of our {ST-BERT-GAT} significantly ($p<0.05$ in $t$-test) outperforms the SOTA baselines and attains 5\% and 2\% improvement on IEMOCAP in terms of ACC and F1 metrics, respectively, while achieving 10\% and 8\% improvement on MELD in terms of ACC and F1, respectively.  By examining more details, we notice that  
\begin{compactitem}
\item ST-BERT outperforms {TL-ERC} (applying BERT), {DRNN$\dag$} (using the RoBERTa features), and {CESTa} (applying Transformer), which exhibits the superiority of our model structure rather than simply using the pre-training models. 
\item ST-BERT also outperforms models using the succeeding context, i.e., {DRNN}, {DRNN$\dag$}, {DGCN}, and {CESTa}.  This implies that our model is more practical in real conversations.
\end{compactitem}
\item ST-BERT also attains the best scores in many individual emotion types, especially ``neutral".  The type of ``neutral" is hard to distinguish because the ``neutral" utterances often contain emotional words.  For example, the word ``excited" colored in blue of Table~\ref{tab:case} may mislead the recognition.  On the contrary, if the model can distinguish the emotional influence among interlocutors, it could easily discover the ``frustrated" emotion from the single word utterance ``Yeah." because most of the time, the emotions of ``frustrated" and ``excited" do not appear simultaneously in a conversation.  The ST-BERT can distinguish the emotional influence through independent temporal modeling and different types of fusion. 
\item By comparing different fusion mechanisms in ST-BERT, we notice that the gate mechanism outperforms others.  Surprisingly, the attention mechanism performs worse than the one by direct concatenation.  We conjecture that the linear combination may not  sufficient absorb the two representations defined in Eq.~(\ref{eq:ST-BERT-att}).  Though the attention mechanism in the Transformer can learn the complex interactions between interlocutors, the hierarchical structure does not provide more information gain than the Flat structure. 
\end{compactitem}

\begin{table}[htp]
\small
\centering
  \caption{Ablation study on IEMOCAP.}
  \label{tab:ab}
  \begin{tabular}{cccc|c}
    \toprule
    target & intra-context & inter-context & pre-train & Avg.(w) F1 \\
    \midrule
    $\surd$       & $\times$      & $\times$      & $\times$  & 49.2    \\
    $\surd$       & $\surd$       & $\times$      & $\times$  & 56.2    \\
    $\surd$       & $\times$      & $\surd$       & $\times$  & 55.5    \\
    $\surd$       & $\surd$       & $\surd$       & $\times$  & 60.8    \\
    \midrule
    $\surd$       & $\times$      & $\times$     & $\surd$   & 55.6    \\
    $\surd$       & $\surd$       & $\times$     & $\surd$   & 63.5    \\
    $\surd$       & $\times$      & $\surd$      & $\surd$   & 61.5    \\
    $\surd$       & $\surd$       & $\surd$      & $\surd$   & 68.4    \\
    \bottomrule
\end{tabular}
\end{table}

\subsection{Ablation Study}
Table~\ref{tab:ab} shows the ablation study on IEMOCAP to test the effect of the components in ST-BERT, including target utterance, intra-context, inter-context, and pre-training.  The first four cases are based on BERT without pre-training, i.e., randomly initializing the parameters. The first case is to test the vanilla BERT$_{\tiny\mbox{Base}}$ without including any context information. The following three cases are to test ST-BERT with intra-context, ST-BERT with inter-context, and ST-BERT with both intra-context and inter-context, respectively.  Similarly, the last four cases are to test the above four cases by the pre-training BERT$_{\tiny\mbox{Base}}$ initialization.  The average F1 scores reported in Table~\ref{tab:ab} clearly show that
\begin{compactitem}
\item Without pre-training, our ST-BERT performs even worse than some baselines in Table~\ref{tab:rs}.  We conjecture that because IEMOCAP is a small dataset, the deep structure of BERT cannot be well-trained in this dataset.
\item Though the models perform poorly when applying randomly initialization, our ST-BERT still beats {TL-ERC} (applying pre-trained BERT). This implies that ST-BERT indeed sufficiently capture emotion dynamics to improve the model performance.
\item When we apply pre-trained BERT on our models, we can obtain significant improvement in all four compared cases. Especially, in ST-BERT, we attain 68.4 F1 score and outperforms {DRNN$\dag$*}, which has applied more advanced pre-training techniques, e.g., RoBERTa.  By comparing other cases, we observe that by including intra-context, more gains can be obtained than by including inter-context.
\end{compactitem}

\begin{figure}[htp]
  \centering
  \subfigure[{BERT$_{\tiny\mbox{Base}}$}] {  
  \label{bs_rs}
  \includegraphics[width=0.445\columnwidth]{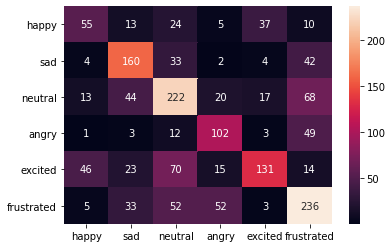}  
  } 
  \subfigure[F-BERT] {  
  \label{fs_rs}
  \includegraphics[width=0.445\columnwidth]{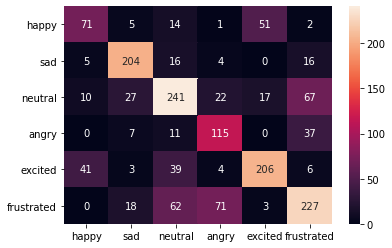}  
  }     
  \subfigure[{H-BERT}]{
    \label{hs_rs}
  \includegraphics[width=0.445\columnwidth]{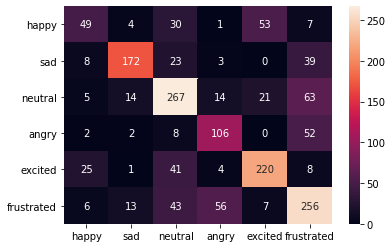}  
  }
  \subfigure[{ST-BERT}]{
    \label{sts_rs}
  \includegraphics[width=0.445\columnwidth]{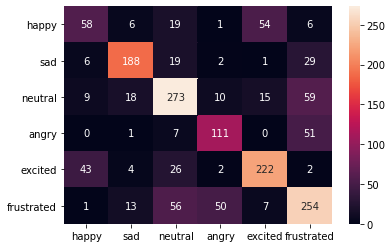} 
  }
  \caption{Heat map of confusion matrix of four BERT-based models.\label{fig:cm}}
\end{figure}

\subsection{Case Study}
We depict the confusion matrices in the form of heat maps to better understand our proposed BERT-based models. The confusion matrices are based on the results of BERT$_{\tiny\mbox{Base}}$, F-BERT, H-BERT, and ST-BERT on IEMOCAP as shown in Fig.~\ref{fig:cm}.  Notice that the types of ``happy", ``excited", and ``neutral" comprise more  easy-to-confuse cases in the positive emotion group while the types of ``sad", ``angry", ``frustrated", and ``neutral"  comprises more easy-to-confuse cases in the negative emotion group.  The ``neutral" type belongs to both groups and is thus particularly hard for recognition.  However, our proposed F-BERT, H-BERT, and ST-BERT perform better than BERT$_{\tiny\mbox{Base}}$ in recognizing ``neutral".  By comparing Fig.~\ref{hs_rs} and Fig.~\ref{sts_rs}, we can notice that ST-BERT exhibits more power in distinguishing the negative emotions and obtain higher performance. 

Table~\ref{tab:case} illustrates a conversation snippet classified by our proposed BERT-based models. There are two challenges in predicting the four utterances.  One is to predict the emotion of a short utterance at the second turn.  The other is to predict the emotion of an utterance with the misleading word ``excited" at the fourth turn.  The results show that ST-BERT correctly predicts the emotion in all four  utterances.  H-BERT cannot distinguish the emotion when there is a misleading word.  Meanwhile, F-BERT fails in predicting both cases.

\begin{table}[htp]
\small
\centering
\caption{A conversation snippet classified by F-BERT (F), H-BERT (H), and ST-BERT (ST). GT stands for ground truth of the emotion types in neutral (Neu.), frustrated (Fru.), and excited (Exc.) .  \label{tab:case}}
\begin{tabular}{|m{6pt}|m{0.7cm}|m{3.4cm}|m{8pt}m{8pt}m{8pt}m{12pt}|}
\hline
&&&&&&\\[-1ex]
 &  A &  B & GT & F  & H    & ST    \\[1.25ex]
\hline\hline
&&&&&&\\[-1.25ex]
\#1    &                        & There's like mystery here, there's magic. It's like a little bit of the unexplainable. I just can't see how you're not interested.                    & Neu.     & Neu. & Neu.    & Neu.    \\[3.5ex]
\hline\hline
&&&&&&\\[-0.75ex]
\#2  & Yeah.                  &                                                                                                                                                       & Fru.   & \red{Neu.} & Fru. & Fru. \\[1.75ex]
\hline\hline
&&&&&&\\[-1.25ex]
\#3  &                        & God, I don't get it. you know the first time we came here, you said it was the best night of your life?                                               & Neu.     & Neu. & Neu.    & Neu.    \\[3.5ex]
\hline\hline
&&&&&&\\[-1.25ex]
\#4  &                        & And last year, I remember distinctly you said, you were so \blue{excited} to get here that you don't remember you stubbed your toe until we were in the car. & Neu.      & \red{Exc.} & \red{Exc.}    & Neu.    \\[4.5ex]
\hline
\end{tabular}
\end{table}

\section{Conclusion}
In this paper, we develop a series of BERT-based models to model emotion dynamics in conversation.  Two basics, a flat-structured and a hierarchically-structured BERT, are proposed to model the preceding utterance information and the direct dependencies in intra-interlocutors and inter-interlocutors.  More importantly, a  spatial-temporal-structured BERT is proposed to specifically distinguish emotional influence among interlocutors, so that we can effectively capture the emotion dynamics.  We conduct extensive experiments on two popular ERC datasets and demonstrate that our proposed BERT-based models can significantly outperform the SOTA baselines.  Detailed ablation study and case study have provided to verify our observations.

\bibliographystyle{plain}
\bibliography{ref}

\begin{thebibliography}{10}

\bibitem{abdul2017emonet}
Muhammad Abdul-Mageed and Lyle Ungar.
\newblock Emonet: Fine-grained emotion detection with gated recurrent neural
  networks.
\newblock In {\em ACL}, pages 718--728, 2017.

\bibitem{arevalo2020gated}
John Arevalo, Thamar Solorio, Manuel Montes-y Gomez, and Fabio~A Gonz{\'a}lez.
\newblock Gated multimodal networks.
\newblock {\em Neural Computing and Applications}, pages 1--20, 2020.

\bibitem{busso2008iemocap}
Carlos Busso, Murtaza Bulut, and Chi-Chun Lee.
\newblock Iemocap: Interactive emotional dyadic motion capture database.
\newblock {\em Language resources and evaluation}, 42(4):335, 2008.

\bibitem{cho2014learning}
Kyunghyun Cho and Bart van Merri{\"e}nboer.
\newblock Learning phrase representations using rnn encoder-decoder for
  statistical machine translation.
\newblock In {\em EMNLP}, 2014.

\bibitem{Clark2020ELECTRA}
Kevin Clark, Minh-Thang Luong, Quoc~V. Le, and Christopher~D. Manning.
\newblock Electra: Pre-training text encoders as discriminators rather than
  generators.
\newblock In {\em ICLR}, 2020.

\bibitem{colneric2018emotion}
Niko Colneri{\^c} and Janez Demsar.
\newblock Emotion recognition on twitter: Comparative study and training a
  unison model.
\newblock {\em IEEE transactions on affective computing}, 2018.

\bibitem{devlin2019bert}
Jacob Devlin, Ming-Wei Chang, Lee, and Toutanova.
\newblock Bert: Pre-training of deep bidirectional transformers for language
  understanding.
\newblock In {\em NAACL}, pages 4171--4186, 2019.

\bibitem{gao2021advances}
Chongming Gao, Wenqiang Lei, Xiangnan He, Maarten de~Rijke, and Tat-Seng Chua.
\newblock Advances and challenges in conversational recommender systems: A
  survey.
\newblock {\em arXiv preprint arXiv:2101.09459}, 2021.

\bibitem{GaoGL19}
Jianfeng Gao, Michel Galley, and Lihong Li.
\newblock Neural approaches to conversational {AI}.
\newblock {\em Foundations and Trends in Information Retrieval},
  13(2-3):127--298, 2019.

\bibitem{ghosal2020utterance}
Deepanway Ghosal, Navonil Majumder, Rada Mihalcea, and Soujanya Poria.
\newblock Utterance-level dialogue understanding: An empirical study.
\newblock {\em arXiv preprint arXiv:2009.13902}, 2020.

\bibitem{ghosal-etal-2019-dialoguegcn}
Deepanway Ghosal, Navonil Majumder, and Soujanya Poria.
\newblock {D}ialogue{GCN}: A graph convolutional neural network for emotion
  recognition in conversation.
\newblock In {\em EMNLP}, 2019.

\bibitem{hazarika2018icon}
Devamanyu Hazarika, Soujanya Poria, and Rada Mihalcea.
\newblock Icon: interactive conversational memory network for multimodal
  emotion detection.
\newblock In {\em EMNLP}, pages 2594--2604, 2018.

\bibitem{hazarika2018conversational}
Devamanyu Hazarika, Soujanya Poria, and Amir Zadeh.
\newblock Conversational memory network for emotion recognition in dyadic
  dialogue videos.
\newblock In {\em NAACL}, 2018.

\bibitem{hazarika2019emotion}
Devamanyu Hazarika, Soujanya Poria, Roger Zimmermann, and Rada Mihalcea.
\newblock Emotion recognition in conversations with transfer learning from
  generative conversation modeling.
\newblock {\em arXiv preprint arXiv:1910.04980}, 2019.

\bibitem{hsu2018emotionlines}
Chao-Chun Hsu, Sheng-Yeh Chen, Chuan-Chun Kuo, Ting-Hao Huang, and Lun-Wei Ku.
\newblock Emotionlines: An emotion corpus of multi-party conversations.
\newblock In {\em ICLRE}, 2018.

\bibitem{DBLP:journals/tois/HuangZG20}
Minlie Huang, Xiaoyan Zhu, and Jianfeng Gao.
\newblock Challenges in building intelligent open-domain dialog systems.
\newblock {\em {ACM} Trans. Inf. Syst.}, 38(3):21:1--21:32, 2020.

\bibitem{DBLP:conf/aaai/JiaoLK20}
Wenxiang Jiao, Michael~R. Lyu, and Irwin King.
\newblock Real-time emotion recognition via attention gated hierarchical memory
  network.
\newblock In {\em {AAAI}}, pages 8002--8009. {AAAI} Press, 2020.

\bibitem{DBLP:conf/naacl/JiaoYKL19}
Wenxiang Jiao, Haiqin Yang, Irwin King, and Michael~R. Lyu.
\newblock Higru: Hierarchical gated recurrent units for utterance-level emotion
  recognition.
\newblock In Jill Burstein, Christy Doran, and Thamar Solorio, editors, {\em
  {NAACL-HLT}}, pages 397--406. Association for Computational Linguistics,
  2019.

\bibitem{kratzwald2018decision}
Bernhard Kratzwald, Suzana Ilic, Mathias Kraus, Stefan Feuerriegel, and Helmut
  Prendinger.
\newblock Decision support with text-based emotion recognition: Deep learning
  for affective computing.
\newblock {\em arXiv preprint arXiv:1803.06397}, 2018.

\bibitem{kuppens2010emotional}
Peter Kuppens, Nicholas~B Allen, and Lisa~B Sheeber.
\newblock Emotional inertia and psychological maladjustment.
\newblock {\em Psychological science}, 21(7):984--991, 2010.

\bibitem{lei2020conversational}
Wenqiang Lei, Xiangnan He, Maarten de~Rijke, and Tat-Seng Chua.
\newblock Conversational recommendation: Formulation, methods, and evaluation.
\newblock In {\em Proceedings of the 43rd International ACM SIGIR Conference on
  Research and Development in Information Retrieval}, pages 2425--2428, 2020.

\bibitem{lei2018sequicity}
Wenqiang Lei, Xisen Jin, Min-Yen Kan, Zhaochun Ren, Xiangnan He, and Dawei Yin.
\newblock Sequicity: Simplifying task-oriented dialogue systems with single
  sequence-to-sequence architectures.
\newblock In {\em Proceedings of the 56th Annual Meeting of the Association for
  Computational Linguistics (Volume 1: Long Papers)}, pages 1437--1447, 2018.

\bibitem{lei2021have}
Wenqiang Lei, Yisong Miao, Runpeng Xie, Bonnie Webber, Meichun Liu, Tat-Seng
  Chua, and Nancy~F Chen.
\newblock Have we solved the hard problem? it’s not easy! contextual lexical
  contrast as a means to probe neural coherence.
\newblock In {\em Proceedings of the AAAI Conference on Artificial
  Intelligence}, 2021.

\bibitem{li2020bieru}
Wei Li, Wei Shao, Shaoxiong Ji, and Erik Cambria.
\newblock Bieru: Bidirectional emotional recurrent unit for conversational
  sentiment analysis, 2020.

\bibitem{liu2019roberta}
Yinhan Liu, Myle Ott, Naman Goyal, Jingfei Du, Mandar Joshi, and Danqi Chen.
\newblock Roberta: A robustly optimized bert pretraining approach.
\newblock {\em arXiv preprint arXiv:1907.11692}, 2019.

\bibitem{loshchilov2018fixing}
I~Loshchilov and F~Hutter.
\newblock Fixing weight decay regularization in adam, corr, abs/1711.05101.
\newblock In {\em Proceedings of the ICLR 2018 Conference}, volume~30, 2018.

\bibitem{DBLP:journals/inffus/MaNXC20}
Yukun Ma, Khanh~Linh Nguyen, Frank~Z. Xing, and Erik Cambria.
\newblock A survey on empathetic dialogue systems.
\newblock {\em Inf. Fusion}, 64:50--70, 2020.

\bibitem{majumder2019dialoguernn}
Navonil Majumder, Soujanya Poria, and Devamanyu Hazarika.
\newblock Dialoguernn: An attentive rnn for emotion detection in conversations.
\newblock In {\em AAAI}, 2019.

\bibitem{mao2020dialoguetrm}
Yuzhao Mao, Qi~Sun, Guang Liu, Xiaojie Wang, Weiguo Gao, Xuan Li, and Jianping
  Shen.
\newblock Dialoguetrm: Exploring the intra- and inter-modal emotional behaviors
  in the conversation, 2020.

\bibitem{mohammad2010emotions}
Saif~M Mohammad and Peter~D Turney.
\newblock Emotions evoked by common words and phrases: Using mechanical turk to
  create an emotion lexicon.
\newblock In {\em NAACL workshop}, 2010.

\bibitem{morris2000emotions}
Michael~W Morris and Dacher Keltner.
\newblock How emotions work: The social functions of emotional expression in
  negotiations.
\newblock {\em Research in organizational behavior}, 22:1--50, 2000.

\bibitem{peters2017semi}
Matthew Peters, Waleed Ammar, Chandra Bhagavatula, and Russell Power.
\newblock Semi-supervised sequence tagging with bidirectional language models.
\newblock In {\em ACL}, pages 1756--1765, July 2017.

\bibitem{peters2018ELMo}
Matthew Peters, Mark Neumann, Mohit Iyyer, Matt Gardner, Christopher Clark, and
  Kenton Lee.
\newblock Deep contextualized word representations.
\newblock In {\em NAACL}, pages 2227--2237, June 2018.

\bibitem{picard2010affective}
Rosalind~W Picard.
\newblock Affective computing: from laughter to ieee.
\newblock {\em IEEE Transactions on Affective Computing}, 1(1):11--17, 2010.

\bibitem{poria2017context}
Soujanya Poria, Erik Cambria, Devamanyu Hazarika, Navonil Majumder, and Amir
  Zadeh.
\newblock Context-dependent sentiment analysis in user-generated videos.
\newblock In {\em ACL}, pages 873--883, 2017.

\bibitem{poria2019meld}
Soujanya Poria, Devamanyu Hazarika, and Navonil Majumder.
\newblock Meld: A multimodal multi-party dataset for emotion recognition in
  conversations.
\newblock In {\em ACL}, pages 527--536, 2019.

\bibitem{poria2019emotion}
Soujanya Poria, Navonil Majumder, Rada Mihalcea, and Eduard Hovy.
\newblock Emotion recognition in conversation: Research challenges, datasets,
  and recent advances.
\newblock {\em IEEE Access}, 7, 2019.

\bibitem{radford2019language}
Alec Radford, Jeffrey Wu, Rewon Child, David Luan, Dario Amodei, and Ilya
  Sutskever.
\newblock Language models are unsupervised multitask learners.
\newblock {\em OpenAI Blog}, 1(8):9, 2019.

\bibitem{serban2016building}
Iulian~V. Serban, Alessandro Sordoni, Yoshua Bengio, Aaron Courville, and
  Joelle Pineau.
\newblock Building end-to-end dialogue systems using generative hierarchical
  neural network models, 2016.

\bibitem{shaheen2014emotion}
Shadi Shaheen, Wassim El-Hajj, Hazem Hajj, and Shady Elbassuoni.
\newblock Emotion recognition from text based on automatically generated rules.
\newblock In {\em IEEE ICDM Workshop}, 2014.

\bibitem{tsai2019multimodal}
Yao-Hung~Hubert Tsai, Shaojie Bai, Paul~Pu Liang, J~Zico Kolter, and
  Louis-Philippe Morency.
\newblock Multimodal transformer for unaligned multimodal language sequences.
\newblock In {\em ACL}, 2019.

\bibitem{tsai2019learning}
Yao-Hung~Hubert Tsai, Paul~Pu Liang, Amir Zadeh, Louis-Philippe Morency, and
  Ruslan Salakhutdinov.
\newblock Learning factorized multimodal representations.
\newblock In {\em ICLR}, 2019.

\bibitem{tzirakis2017end}
Panagiotis Tzirakis, George Trigeorgis, and Mihalis~A Nicolaou.
\newblock End-to-end multimodal emotion recognition using deep neural networks.
\newblock {\em IEEE Signal Processing}, 2017.

\bibitem{vaswani2017attention}
Ashish Vaswani, Noam Shazeer, Niki Parmar, Jakob Uszkoreit, Llion Jones,
  Aidan~N Gomez, {\L}ukasz Kaiser, and Illia Polosukhin.
\newblock Attention is all you need.
\newblock In {\em NIPS}, pages 5998--6008, 2017.

\bibitem{wang2015constructing}
Feng Wang, Xiaoyan Li, Wenqiang Lei, Chen Huang, Min Yin, and Ting-Chuen Pong.
\newblock Constructing learning maps for lecture videos by exploring wikipedia
  knowledge.
\newblock In {\em Pacific Rim Conference on Multimedia}, pages 559--569.
  Springer, 2015.

\bibitem{wang2020contextualized}
Yan Wang, Jiayu Zhang, Jun Ma, Shaojun Wang, and Jing Xiao.
\newblock Contextualized emotion recognition in conversation as sequence
  tagging.
\newblock In {\em ACL}, pages 186--195, 2020.

\bibitem{WuSCLNMKCGMKSJL16}
Yonghui Wu, Mike Schuster, Zhifeng Chen, and Quoc~V. Le.
\newblock Google's neural machine translation system: Bridging the gap between
  human and machine translation.
\newblock {\em CoRR}, 2016.

\bibitem{zhang2020generalized}
Yao Zhang, Xu~Zhang, Jun Wang, Hongru Liang, Wenqiang Lei, Zhe Sun, Adam
  Jatowt, and Zhenglu Yang.
\newblock Generalized relation learning with semantic correlation awareness for
  link prediction.
\newblock {\em arXiv preprint arXiv:2012.11957}, 2020.

\end{thebibliography}
\end{document}